\documentclass[fleqn,10pt,x11names]{wlscirep}

\usepackage[superscript,biblabel]{cite}
\usepackage{comment}

\usepackage[utf8]{inputenc}
\usepackage{multicol}
\usepackage{amssymb}
\usepackage{hyperref}
\usepackage{amsmath}
\usepackage{graphicx}
\usepackage[normalem]{ulem}
\usepackage{makecell}
\usepackage{csquotes}
\usepackage{soul}
\usepackage{subfig}
\usepackage{longtable}
\usepackage{lscape}

\usepackage{booktabs}
\usepackage{pdflscape}
\usepackage{longtable}
\usepackage{amssymb}
\usepackage{pdfpages}

\usepackage{graphicx}
\usepackage{tabularx,booktabs}
\let\checkmark\undefined
\usepackage{dingbat}
\usepackage{diagbox}
\usepackage{pifont}
\newcolumntype{C}{>{\centering\arraybackslash}X}
\newcommand{\xmark}{\ding{55}}%

\usepackage{titlesec}

\title{Recent Methodological Advances in Federated Learning for Healthcare}
\author[1]{Fan Zhang}
\author[1]{Daniel Kreuter}
\author[1]{Yichen Chen}
\author[1,2]{S{\"{o}}ren Dittmer}
\author[1]{Samuel Tull}
\author[3]{Tolou Shadbahr}
\author[4]{BloodCounts! Collaboration}
\author[5]{Jacobus Preller}
\author[6]{James H.F. Rudd}
\author[7]{John A.D. Aston}
\author[1]{Carola-Bibiane Sch{\"{o}}nlieb}
\author[8]{Nicholas Gleadall}
\author[1,6,*]{Michael Roberts}

\affil[1]{Department of Applied Mathematics and Theoretical Physics, University of Cambridge, Cambridge, UK}
\affil[2]{ZeTeM, University of Bremen, Bremen, Germany}
\affil[3]{Research Program in Systems Oncology, Faculty of Medicine, University of Helsinki, Helsinki, Finland}
\affil[4]{A list of authors and their affiliations appears at the end of the paper.}
\affil[5]{Addenbrooke’s Hospital, Cambridge University Hospitals NHS Trust, Cambridge, UK.}
\affil[6]{Department of Medicine, University of Cambridge, Cambridge, UK}
\affil[7]{Department of Pure Mathematics and Mathematical Statistics, University of Cambridge, Cambridge, UK}
\affil[8]{Department of Haematology, University of Cambridge, Cambridge, UK}

\affil[*]{corresponding author: mr808@cam.ac.uk}

\begin{abstract}
For healthcare datasets, it is often not possible to combine data samples from multiple sites due to ethical, privacy or logistical concerns. Federated learning allows for the utilisation of powerful machine learning algorithms without requiring the pooling of data. Healthcare data has many simultaneous challenges which require new methodologies to address, such as highly-siloed data, class imbalance, missing data, distribution shifts and non-standardised variables. Federated learning adds significant methodological complexity to conventional centralised machine learning, requiring distributed optimisation, communication between nodes, aggregation of models and redistribution of models. In this systematic review, we consider all papers on Scopus that were published between January 2015 and February 2023 and which describe new federated learning methodologies for addressing challenges with healthcare data. We performed a detailed review of the 89 papers which fulfilled these criteria. Significant systemic issues were identified throughout the literature which compromise the methodologies in many of the papers reviewed. We give detailed recommendations to help improve the quality of the methodology development for federated learning in healthcare. 
\end{abstract}

\usepackage{xcolor}

\begin{document}

\maketitle

\begin{multicols}{2}

\section{Introduction}
\label{sec:intro}

Healthcare data is abundant, representing approximately 30\% of the entire global data volume \cite{thomason2021big}, and is becoming increasingly available to researchers to allow for such interrogation as trend analysis, pattern recognition and predictive modelling. This is helped primarily by the increased adoption of electronic health record (EHR) systems in hospitals, with most UK NHS Trusts currently using one and all expected to have one by 2025 \cite{health_plan}. In parallel, there has been a revolution in the capabilities of machine learning (ML) methods, allowing for the efficient analysis of high-dimensional clinical and imaging data.

There are different types and formats of healthcare data, including text from medical notes, imaging data, medical device outputs, wearable signals data and genomic data. These are usually stored in distinct silos, with EHR data often in a database structure, imaging in a picture archiving and communication system (PACS) and medical device/wearable data stored locally.
Although there are some well-documented challenges to reproducibility \cite{kapoor2023leakage, dittmer2023navigating}, ML methods have shown great utility for performing both single and multiple modality modelling of healthcare data \cite{defauwClinicallyApplicableDeep2018, isenseeNnUNetSelfconfiguringMethod2021}.

To create high-quality models which generalise across different data sources, it is most common to pool datasets from different locations and train using the combined dataset. However, this provides a serious challenge in the healthcare setting as there are ethical and privacy concerns regarding the transfer of clinical data outside the hospital environment. Additionally, there are logistical issues to ensure data security is maintained in the transfer of such large-scale healthcare data. Finally, each hospital might have its own rules on data transfer and sharing, making research across more than one hospital problematic.

Federated learning (FL) offers a solution to the latter problem by permitting data to remain locally at each of the different hospital sites, with only the ML model being transferred between them. An FL network can be either decentralised or have a central aggregator which communicates with all nodes. In the decentralised setting, a model is trained at individual sites. The updated model is passed around to other sites in the network in a defined order, who then initialise from the model, train and pass it on again. In the centralised aggregator scenario, an ML model is trained at each site and information about the final model is transferred from all nodes to the aggregator. The aggregator summarises the model updates from each site and generates a new global model which is then redistributed to all sites for the training process to start again locally. This continues until a pre-defined convergence criteria is met for the global model. 

\begin{figure*}
    \centering
\subfloat[HFL]{
   \includegraphics[height=4cm,width=5cm]{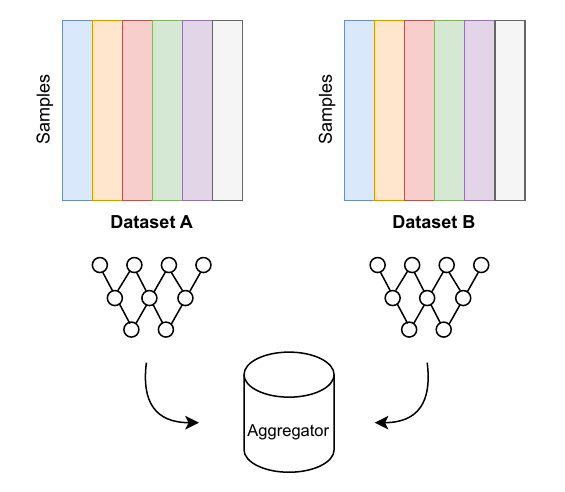}}\quad
\subfloat[VFL]{
   \includegraphics[height=4cm,width=5cm]{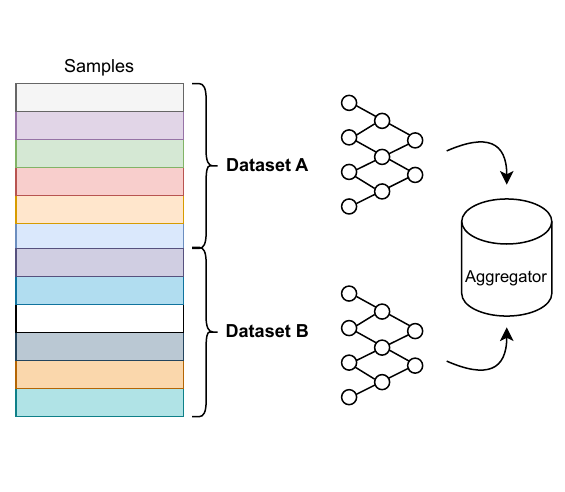}}\quad
\subfloat[FTL]{
   \includegraphics[height=4cm,width=5cm]{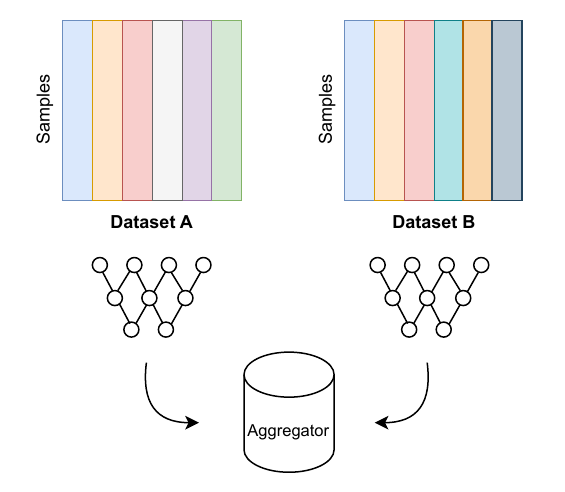}}
    \caption{Schematic illustrating the feature and sample distributions for Horizontal Federated Learning (HFL), Vertical Federated Learning (VFL) and Federated Transfer Learning (FTL).}
    \label{fig:FLclasses}
\end{figure*}

FL methods can be broadly categorised into three groups: horizontal, vertical, and transfer \cite{yang2019} (see Figure~\ref{fig:FLclasses}). For horizontal FL (HFL) methods, each site holds the same features for different data samples, whereas for vertical FL (VFL), the sites hold different features related to the same samples. For federated transfer learning (FTL), each site has different feature sets that are related to different samples \cite{yang2019}. Each of these has high relevance for healthcare data, where HFL is akin to learning across different hospitals for common variables and VFL allows for linking of different data silos for, e.g. imaging, electronic health record (EHR) and genomic data. FTL applies most to the real-world clinical environment where different hospitals collect different variables on different people, often dependent on their local clinical protocols.

Healthcare data exhibits many issues that require special consideration and adjustments to FL methodologies to address. Crucially, in a network of hospitals, each hospital can potentially serve a fundamentally different patient population. This bias can lead to downstream modelling issues if there are, for example, different disease prevalences at different sites or significant differences in the patient numbers at each location \cite{rolandDomainShiftsMachine2022}. Each hospital may also follow different clinical practices, leading to differences in data collection and consequently issues such as missing data \cite{haneuseAssessingMissingData2021} and non-standardised variable name mappings \cite{overhageValidationCommonData2012}. Privacy is a top concern for all users of healthcare data and it is critical that FL methods for healthcare are considerate of this when transferring model parameters between the sites or to the aggregator \cite{nassValueImportanceHealth2009}. Finally, we highlight that hospital environments tend not to be equipped with high-performance computational environments, so FL methods must factor in the computational cost of training, transferring and evaluating models.

There is significant literature describing FL methods applied to healthcare data with applications such as breast cancer diagnosis \cite{221, 61, 224, 344, 416, 14, 108}, COVID-19 detection  \cite{10, 27, 51, 29, 377, 248, 139, 194, 202, 205, 283, 47}, length of hospital stay prediction \cite{89} and depression diagnosis \cite{185, 381}. Indeed, in this systematic review, we identified 214 such papers between 2015 and 2023. Of these, 125 papers apply existing FL methods off-the-shelf to healthcare datasets and 89 papers describe modifications of the FL methodology to address the challenges unique to healthcare data. It is the latter group that this review focuses on, as we are keen to identify which areas of methodology advancement are receiving the focus and whether there are any systematic pitfalls in the way that these new models are being developed.

\begin{figure*}
    \centering
    \includegraphics[width=0.8\textwidth]{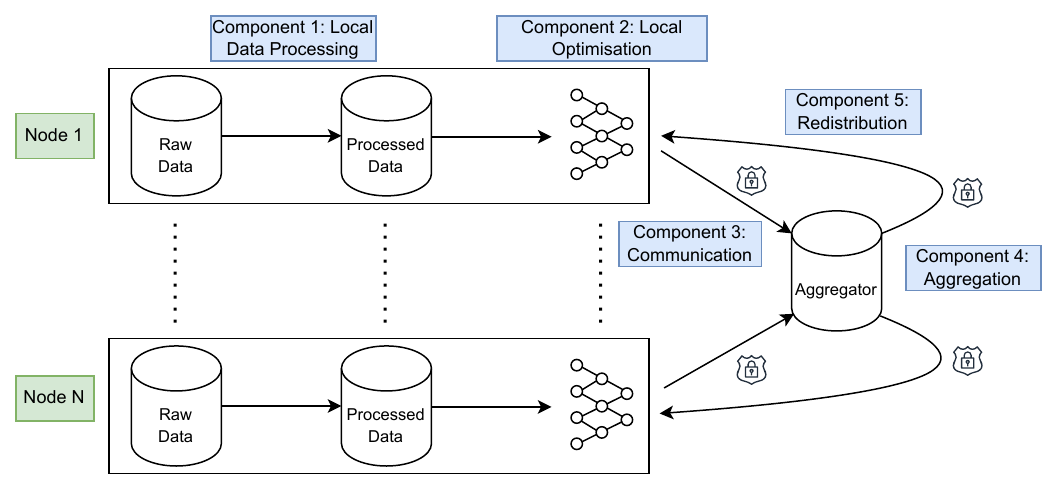}
    \caption{Typical FL workflow with the five components identified which form the basis of our analysis.}
    \label{fig:flflow}
\end{figure*}

There were three systematic reviews relevant to ours that assess FL applications to healthcare \cite{antunes2022federated, prayitno2021systematic, crowson2022systematic} which considered 44, 24 and 13 papers, respectively. Only one study\cite{crowson2022systematic} highlighted issues in approaches described in the papers. In this systematic review, we focus on 89 papers that develop new FL methods and apply them to healthcare. Our analysis splits FL into five distinct components (see Figure~\ref{fig:flflow}) and focuses on the methodological approaches for each. We carefully review each paper with respect to these components to identify advances in the methodologies and any systemic issues. In the light of the issues we identify, we also make recommendations for each component for researchers and practitioners to encourage more reproducible high-quality FL methodologies to be developed for healthcare applications.

\begin{figure*}
    \centering
    \includegraphics[width=0.8\textwidth]{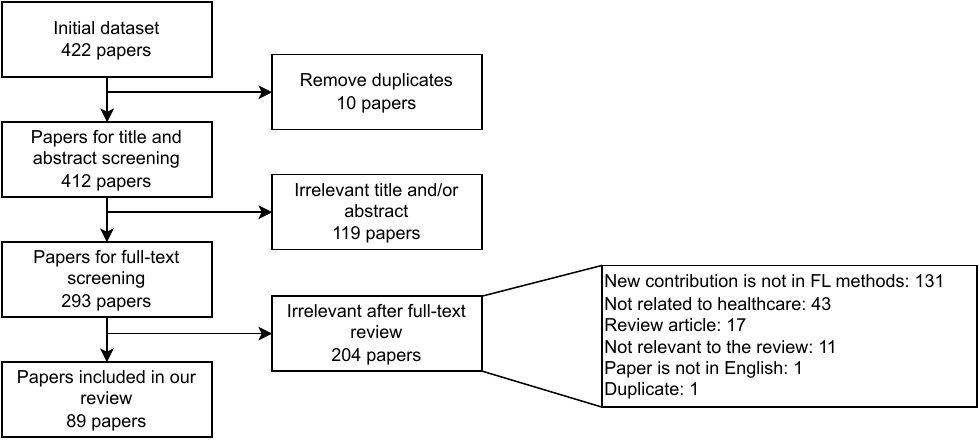}
    \caption{PRISMA flowchart for our systematic review highlighting the reasons for exclusion of manuscripts at different stages.}
    \label{fig:prisma}
\end{figure*}

\section{Results}
\label{sec:results}

This section will first give a general overview of the reviewed studies and their methodological advances.
Each subsequent section we carefully consider the five distinct components of FL depicted in Figure~\ref{fig:flflow} with respect to the reviewed studies and summarise the findings in Table~\ref{Tab:1}. \\

\noindent
\emph{Study selection.} The initial search identified 422 papers that met the search criteria (see Methods section). After eliminating 10 duplicate papers and filtering to only abstracts and titles focusing on a new approach of FL in healthcare, we retained 293 papers for full-text screening. In this systematic review, we focus on the 89 papers that were relevant to our review question, namely those that introduce a new methodology for applying FL in the context of healthcare.\\

\noindent
\emph{Methodology advances.} We consider the methodological contributions made to the different FL components identified in Figure~\ref{fig:flflow}. The majority of papers (68/89) contribute to a single component, with 18/89 contributing to two and 1/89 contributing to three. Most papers focus on improving the aggregation component (37/89), followed by the communication (35/89) and local optimisation (26/89) components. The local data processing component is improved in 5/89 papers and model redistribution in 4/89. \\

\noindent
\emph{Types of FL.}
HFL was the most popular among the three approaches, with 80/89 studies exclusively considering it \cite{9, 10, 12, 13, 14, 16, 20, 27, 29, 30, 31, 35, 38, 51, 58, 60, 61, 64, 76, 77, 83, 89, 95, 97, 101, 108, 111, 112, 115, 136, 139, 142, 143, 150, 154, 155, 165, 167, 180, 185, 191, 192, 194, 202, 205, 206, 213, 215, 221, 222, 224, 243, 245, 248, 257, 265, 267, 270, 272, 283, 300, 318, 328, 332, 333, 343, 344, 346, 355, 365, 371, 375, 377, 378, 379, 381, 385, 389, 391, 416}. In contrast, VFL~\cite{106, 137, 169, 373, 393} is considered in only five papers and FTL~\cite{47, 107} in four, while two studies \cite{69, 170} considered both HFL and VFL together.  

VFL was used when the joining of features across data sources was discouraged due to privacy or logistical concerns. For example, when pooling imaging in different modalities \cite{137}, different sensors recording data about the same subject \cite{373} or genotype and phenotype data in different sites. FTL was primarily used with clinical data collected under different protocols where the set of available variables differs between nodes.\\

\noindent
\emph{Applications considered.}
The majority of papers (70/89) applied FL to classification problems, %
segmentation was addressed in 7/89 %
reviewed studies and the remainder focused on problems such as anomaly detection \cite{365, 243}, tensor factorisation \cite{371, 270}, feature selection \cite{169} and regression \cite{14, 29, 47, 343, 83, 143, 165}. See Table~\ref{Tab:1} for full details. \\

\noindent
\emph{Use of existing frameworks.}
We find that most papers considered in this review (83/89) develop their own FL framework for FL rather than building on existing platforms such as Flower \cite{beutel2020flower}, TensorFlow Federated \cite{The_TensorFlow_Federated_Authors_TensorFlow_Federated_2018, tensorflow2015-whitepaper}, PySyft \cite{ziller2021pysyft}, FATE \cite{fate_paper} and NVIDIA FLARE \cite{Roth_NVIDIA_FLARE_Federated_2022}. Six papers used existing frameworks, namely FATE \cite{106}, Flower\cite{283}, NVIDIA FLARE \cite{137, 389} and PySyft \cite{61, 267, 265}.\\

\noindent
\emph{Public codebases.}
Only 10/89 studies publicly released their code \cite{13, 29, 35, 248, 385, 142, 224, 76, 14, 12} and no papers shared their trained model.

\subsection{FL Component Analysis}
\label{sec:flcomponents}

We now review each component of the FL pipeline as outlined in Figure~\ref{fig:flflow}.\\

\noindent
\textbf{Component 1: Local Data Processing} \\

\noindent
\emph{Methodological advances.} In 5/89 papers, authors focus on improving the data pre-processing. Of these, 4/5 were motivated by problems of class imbalance, with three \cite{60, 83, 245} using a generative model to create new data samples and one \cite{377} using augmentation to reduce sample size imbalances across nodes. The remaining paper \cite{346} used anonymisation techniques (such as quantization of some continuous features) to enhance the privacy of the raw data samples. \\

\noindent
\emph{Datasets.}
The papers considered a wide range of data types. The most popular sources were imaging data (31/89), sensor data (17/89), and EHR or tabular clinical data (21/89). Other studies considered more niche data sources such as medical devices (7/89), insurance claims data (4/89), and genomic data (2/89).
For the imaging studies, a variety of modalities were considered. Chest X-ray (9/31) \cite{10, 27, 139, 194, 202, 205, 248, 283, 377}, retinal (6/31) \cite{83, 115, 136, 150, 167, 205}, microscopy (5/31) \cite{9, 61, 77, 191, 416}, dermoscopy (4/31) \cite{192, 205, 245, 378}, and magnetic resonance (3/31) \cite{137, 180, 222} imaging constituted the majority. Sensor data was collected from wearable technologies (14/17) \cite{13, 20, 60, 64, 111, 142, 202, 206, 224, 267, 318, 375, 385, 391} and ambient sensors (4/17) \cite{69, 101, 267, 373}.
For papers using EHR or tabular clinical datasets, MIMIC-III \cite{johnson2016mimic} was used in 7/21 \cite{12, 89, 170, 270, 343, 346, 371}, synthetic EHR data in 4/21 \cite{143, 270, 371, 393}, and proprietary data in 2/21 \cite{328, 346}. 
Offline medical devices data was sourced from electrocardiograms (5/7) \cite{38, 95, 106, 107, 379} and electromyograms (2/7) \cite{101, 318}, whilst insurance claims data was sourced from the Centres for Medicare \& Medicaid Services \cite{270, 371}, UnitedHealth Group Clinical Research \cite{29}, and MarketScan Research \cite{29, 31}.
Ten papers also used popular non-medical datasets, including MNIST (10/12) \cite{9, 27, 136, 332, 142, 245, 185, 215, 257, 375}, CIFAR10 (6/12) \cite{27, 136, 215, 245, 333, 375}, Fashion-MNIST (2/12) \cite{9, 248}, and STL10 (1/12) \cite{27} in order to benchmark their proposed method's performance.\\

\noindent
\emph{Outcomes of interest.}
The scope of the applications considered in the papers was very diverse. A large number (51/89) focused on the diagnosis of diseases such as COVID-19 (12/51) \cite{10, 27, 51, 29, 377, 248, 139, 194, 202, 205, 283, 47}, lung cancer (3/51) \cite{9, 221, 58}, breast cancer (7/51) \cite{221, 61, 224, 344, 416, 14, 108},  skin disease (6/51) \cite{150, 191, 192, 205, 245, 378}, eye disease (4/51) \cite{115, 205, 136, 150}, heart disease (5/51) \cite{106, 107, 300, 379, 393}, brain tumour (4/51) \cite{180, 136, 137, 222}, diabetes (2/51) \cite{83, 167}, neurodegenerative disorders (2/51) \cite{318, 373}, Alzheimer’s disease (2/51) \cite{97, 257}, colorectal carcinoma (2/51) \cite{77, 58}, and sepsis \cite{76}. There are also applications varying from using bone imaging to predict age \cite{83}, detection of emotion in speech \cite{154} and identifying depression through social networking interactions and posts \cite{185}. \\

\noindent
\emph{Data pre-processing.}
Pre-processing of datasets was mentioned in only 41/89 papers, being performed either centrally (before data was allocated to the nodes for synthetic FL experiments) or at a node-by-node level.
Commonly used techniques include feature value normalisation \cite{27, 83, 205, 213, 222, 272, 300, 381}, dimensionality reduction \cite{12, 20, 27, 89, 300}, feature engineering through a data-specific transformation \cite{20, 27, 60, 64, 69, 333, 346}, and data filtering \cite{20, 112, 272, 300, 170, 202, 76, 213, 222, 31} for tabular data. For video data, resampling was used \cite{165}, and for imaging data, resizing \cite{77, 83, 150, 378, 115}, intensity windowing \cite{111}, sliding-window \cite{60, 111, 318, 202} and cropping \cite{30, 83, 416, 222} were employed.

Most papers do not mention performing quality or integrity checks on the data before or after pre-processing, exceptions being Gad et al.\cite{13}, where samples with impossible values are excluded (e.g. negative heart rates) or where inconsistencies in feature values are identified, and Shaik et al.\cite{20} which performs Principal Component Analysis to filter out noise. One paper \cite{393} examined genotype data for discrepancies and anomalies to ensure reliability and accuracy. Within the 89 papers reviewed, 131 datasets were considered with the smallest dataset having 116 samples \cite{108} and the largest having 3.7 million \cite{346}. None of the papers applied hashing or encryption to the raw data before training \cite{lee2018privacy}. \\

\noindent
\emph{Imbalances in data.}
There are two potential sources of imbalanced data for FL. Firstly, different nodes in the network can be associated with highly varying sample sizes, and secondly, the prevalence of the outcome variable (class imbalance) can also vary between nodes. In the studies reviewed, imbalances in sample sizes were addressed in only three papers,  which employed downsampling \cite{272} and augmentation \cite{377, 192}.

Imbalance in outcome prevalence was addressed in only four studies. SMOTE \cite{chawla2002smote} oversampling was used in two papers \cite{10, 60} to generate a balanced dataset at each node, and deep reinforcement learning was used by Zhang et al.~\cite{267} to encourage devices with balanced class labels to participate more frequently in the local updates. A weighted random forest, assigning higher weights to the less prevalent classes, was used by Gencturk et al.~\cite{167}. \\

\noindent
\emph{Consistency across nodes.}
Almost all papers performed their FL experiments in controlled environments with no determination for whether datasets or their feature values are comparable across the different nodes. 
The exceptions were Tong et al.~\cite{29}, who computed a surrogate pairwise likelihood function to account for bias between the model parameters from each site and used this to adjust the final model predictions and Chen et al.~\cite{318} who used the aggregator to compare each node's data distribution with a benchmark dataset to estimate the label quality, reflecting the reliability and accuracy of each node's local data labels. In addition, no papers discussed how the dataset used at each node is curated for use in the experiments.\\ 

\noindent
\emph{Missing data.}
Most studies did not highlight whether datasets contained any missing values, with only two considering an imputation method whilst also deleting features with high missingness rates \cite{373, 12}.\\

\noindent
\textbf{Component 2: Local Optimisation}\\

\noindent
\emph{Methodological advances.} In 26/89 papers, authors focus on improving local optimisation. In 6/89 papers, the authors focus on improving the training procedure by using gradient clipping \cite{137}, modifying the activation function to better tolerate data heterogeneity \cite{64}, regularisation of the model by penalising deviations from earlier time-series datapoints \cite{224}, use of contrastive learning for federated model pre-training \cite{12}, use of parameter sharing to reduce model size \cite{391} and by reordering of the data samples to process the most difficult samples at the end of training \cite{343}.
Six papers focus on improving the model architecture to allow for multi-modal data \cite{213} by tuning the architecture to local datasets \cite{344}, using smaller binarised neural networks for resource constrained settings \cite{385}, using extremely randomised trees for privacy preservation \cite{27}, using an extreme learning machine to directly find the model parameters in one iteration \cite{373} and introducing a new method for survival analysis to factor in time-varying covariates \cite{283}.  
In 9/89 papers \cite{150, 378, 14, 60, 191, 333, 95, 170, 248}, the authors decompose the model into parts that are trained on the local nodes and parts trained on the central server. This gives some aspects of the model, which is fine-tuned to local data. In 4/89 papers, authors design approaches to perform federated semi-supervised learning to utilise unlabelled data samples. In one paper, the authors design a highly stratified cross-validation strategy based on confounding factors to overcome distribution differences between local datasets. \\

\noindent
\emph{Model Architectures.}
In the 82/89 studies that described their model, there was a wide range of complexity, ranging from highly intricate and parameterised models to simpler, traditional ML methods.
The most popular choice for model architecture was a convolutional neural network (CNN) with 46/89 studies considering them. 
Most papers (see Table~\ref{Tab:1}) discuss their own custom CNN architectures (27/46), %
whilst many of these papers (19/46) compare several different established architectures such as ResNet (9/40) \cite{136, 378, 30, 61, 248, 83, 115, 194, 191}, DenseNet (3/40) \cite{10, 9, 30}, MobileNet (2/40) \cite{106, 248}, U-Net (4/40) \cite{137, 150, 180, 283}, AlexNet (4/40) \cite{61, 101, 333, 202}, and LeNet (2/40) \cite{101, 202}. 
Recurrent neural networks (RNNs) were also popular, with LSTM and Bi-LSTM backbones found in 11/89 studies \cite{12, 13, 20, 38, 64, 69, 89, 95, 170, 257, 365} and five papers using both CNNs and RNNs \cite{13, 20, 38, 95, 257}.
Multilayer perceptrons (MLPs) were used in 11 studies, with custom architectures for a vanilla MLP employed in 10 \cite{14, 51, 77, 108, 154, 206, 213, 272, 328, 344} and an attention layer incorporated in one \cite{385}. 
Some studies also considered FL with more traditional ML algorithms such as gradient-boosted trees \cite{108, 169, 343}, support vector machines \cite{108, 169, 300, 346}, fuzzy clustering \cite{112}, logistic regression \cite{29, 31, 97, 108, 143, 346, 393} and random forests \cite{108, 167, 169, 224, 355}. The remaining studies either used custom algorithms or focused on other aspects of the FL pipeline.\\

\noindent
\emph{Optimisers.} 
Of the 59 papers which mentioned their optimisation algorithm, stochastic gradient descent (SGD) is the most widely used method, found in 35/59 papers 
while Adam was used in 20/59 papers (see Table~\ref{Tab:1}).
Other methods such as RMSprop \cite{13, 69, 107}, SAGA \cite{136}, Adadelta \cite{377}, Maximum Likelihood Estimation \cite{143}, or Newton's Method \cite{393} were also considered. One paper developed their own optimiser \cite{167}.\\

\noindent
\emph{Training initialisation.} 
Only 68/89 studies mentioned the use of initialisation strategies for their model parameters. Initialisation with random weights was most common (45/67) %
whilst 16 papers %
utilised predetermined parameters (identified by the authors), 7 used pre-trained weights \cite{30, 379, 139, 185, 202, 205, 283} and another used the parameters obtained by the node whose dataset has the highest number of features \cite{112}. See Table~\ref{Tab:1} for details. \\

\noindent
\emph{Dataset partitioning.} 
Of the papers, 47/89 stated that they partitioned their dataset into internal validation or holdout cohorts. Only two papers \cite{9, 12} used both an internal and external holdout cohort.
14 papers used an internal validation dataset to avoid model overfitting, with 13 using an internal validation cohort of between 5\% and 25\% of the dataset \cite{10, 12, 30, 58, 77, 83, 106, 137, 154, 180, 192, 389, 416} and one using an external dataset for validation \cite{47}. 15 studies performed three- \cite{224, 31}, five- \cite{14, 64, 89, 107, 224, 272, 343, 346, 391} or ten-fold \cite{38, 95, 97, 167} cross-validation whilst two papers \cite{60, 213} mentioned the use of cross-validation without specifying the number of folds. In 22 papers \cite{13, 20, 101, 107, 115, 136, 139, 142, 170, 185, 194, 202, 205, 206, 221, 222, 283, 332, 377, 378, 379, 381}, there is a holdout cohort used for model evaluation after a fixed number of rounds of local optimisations. Three papers mentioned the use of test datasets \cite{300, 373} or validation datasets \cite{318} but not their sizes. \\

\noindent
\emph{Hardware.} 
Only 40/89 papers mentioned the hardware used in their study. The computational requirements for model optimisation varied significantly and consequently, the hardware used was highly diverse. Most FL methods required either hardware with GPU(s) attached (26/40), 
or simple CPU machines (15/40). %
Some studies use edge devices such as a Raspberry Pi \cite{12, 107, 165, 365, 377} or a smartphone \cite{101}.\\ 

\noindent
\emph{Privacy preserving optimisation.}
Six papers \cite{27, 77, 97, 243, 272, 333} applied Differential Privacy~\cite{dwork2006calibrating} during the optimisation. This aimed to preserve the privacy of individual samples in the training data. Five papers accomplished this by adding Gaussian noise to the exchanged data, with the sole exception of Hilberger et al.~\cite{77} who used TensorFlow Privacy~\cite{andrewTensorFlowPrivacy2023} instead.\\

\noindent
\emph{Training termination.} 
Criteria for ending the local optimisation are mentioned in 75/89 papers. Most of which (67/75) %
used a fixed number of local training epochs for each node (see Table~\ref{Tab:1}), whilst four papers \cite{16, 64, 143, 391} terminated on convergence but did not provide details on how this was determined. Two papers \cite{180, 373} only required one epoch for training by design. \\

\noindent
\emph{Decentralised training.}
In seven papers \cite{64, 111, 205, 206, 344, 365, 381}, the training of the model itself was not entirely performed on the nodes. Three studies trained using both the nodes and edge devices \cite{344, 365, 205}, and four studies trained small clusters of nodes which then communicate to the aggregator \cite{64, 111, 206, 381}.\\ 

\noindent
\emph{Novel developments.}
One paper removed the need to perform iterations of an algorithm by use of a one-layer MLP, whose weights can be directly solved for \cite{373} and another paper partitioned the model architecture and then trained the computationally intensive part on the central aggregator \cite{333}. Knowledge distillation was used in two studies \cite{13,30} where more powerful ``teacher'' nodes train less powerful ``student'' nodes. One paper introduced a method where all but the batch normalisation layers are exchanged between each node \cite{416}.\\

\noindent
\textbf{Component 3: Communication}\\

\noindent
\emph{Methodological advances.} In 35/89 papers, there are contributions to improve the communication component with some papers describing contributions to several aspects. Primarily, authors focused on improving encryption methods (11/35), aimed to reduce either the amount of data communicated (12/35) or the number of communication rounds (7/35) and also introduced methods for fully decentralised communication (5/35) \cite{222,328,180,205,300}. Encryption tended to focus on methods for sharing secret keys among nodes \cite{332, 76}, encryption mechanisms for the data exchanged \cite{97,155,169,333,355,391,106, 192} and also a technique for perturbing model outputs at each node using a secret key \cite{139}. A reduction in the amount of data communicated is achieved by transferring a subset of the model parameters \cite{107, 270, 38, 111, 202, 206, 283, 416}, or by compressing, masking and quantising of gradients or model outputs before exchange \cite{371, 272, 215, 106}. The number of rounds of communication with the server can be reduced by the inherent design of the model \cite{29,83,89,143}, aggregating based on elapsed time (rather than epochs) \cite{215, 371} and by checking whether the proposed update is beneficial to the network before communicating it \cite{332}. The remaining papers focus on techniques for detecting attacks during communication \cite{243}, developing an authentication system for nodes in the network \cite{265, 16} and systems for client/node management \cite{142, 192}. \\

\noindent
\emph{Data exchanged.}
In the studies we considered, the data most commonly shared with the aggregator were the model weights (35/89) 
and gradients (13/89), see Table~\ref{Tab:1}.
Many papers simply state that they are exchanging ``model parameters'' (20/89) \cite{12, 14, 16, 27, 47, 58, 64, 69, 89, 112, 115, 142, 150, 154, 202, 243, 272, 300, 389, 416} or ``model updates'' (2/89) \cite{10, 318} without specifically stating if or how these relate to weights or gradients. The outputs of the local models were shared in 7/89 \cite{13, 20, 30, 31, 106, 108, 170} of the papers reviewed. Some papers encrypted the gradients before sharing (4/12) \cite{76, 332, 333, 355}, some encrypted the weights (9/35) \cite{60, 61, 97, 155, 185, 221, 215, 222, 391} and three papers perturbed the model parameters before sharing \cite{27, 243, 272, 344}.

Beyond the conventional gradient and weight data shared in FL methods, some papers also shared additional information about the data or the training process or procedure. Metadata such as the model architecture, optimiser, loss function \cite{77}; training time, maximum performance \cite{10, 222}; the initial learning rate, and thresholds used for classification and learning strategy \cite{38} were among the extra information exchanged.\\

\noindent
\emph{Encryption.}
Data was exchanged unencrypted for the majority of the studies we considered (70/89). Most papers that did encrypt the exchanged data primarily used homomorphic encryption methods (13/19) \cite{60, 61, 76, 106, 155, 169, 192, 215, 221, 224, 333, 355, 391} whilst Chen et al.~\cite{332} relied on a symmetric-key algorithm for encrypting the gradients when exchanged. Only three papers \cite{35, 265, 222} discussed encryption of the communication channel between the node and the aggregator rather than direct encryption of the data before exchange. Two papers \cite{97, 185} stated that they encrypted their data without detailing the method used.\\

\noindent
\emph{Privacy-preserving communication.}
The privacy of the models at individual nodes may be maintained by exchanging only a subset of the layers, as found in 12 papers \cite{38, 95, 107, 111, 170, 202, 206, 248, 283, 333, 375, 416}, some intermediate model outputs \cite{373}, or randomly shuffled model outputs \cite{139}.\\

\noindent
\emph{Authentication of nodes.} 
Only two papers \cite{16, 265} ensured that before the exchanged data is accepted by the aggregator, the node is authenticated as part of the approved FL network. The local model parameters were shared, along with a ring signature, to prove the data originated from the node.\\

\noindent
\emph{Fully decentralised communication.} 
There were nine papers describing fully decentralised FL methods \cite{30, 101, 112, 180, 205, 222, 300, 328, 371}. These require communication between nodes rather than a central aggregator, using gradient and weight updates of neighbour nodes to update the local model. Knowledge distillation between the local nodes is employed in two studies \cite{13, 30} where more powerful nodes behave as teachers to the less powerful student nodes. None of the papers considered the significance of the sequence of updates in the fully decentralised scenario.\\

\noindent
\emph{Optimising efficiency.} 
Two papers considered approaches to reduce the number of rounds of communication with the aggregator by sharing embeddings of the data with the aggregator (which then trains the model) \cite{373} and by modelling the pairwise relationships between the data at each node directly \cite{29}, both of which allow for only a single round of communication.
Communication time is optimised in one paper \cite{205}, where an optimal ring structure between nodes is obtained by solving a version of the travelling salesman problem.
Additionally, eight papers focused on reducing the amount of data transferred between the nodes and the aggregator at each iteration. This can be accomplished by sharing only a subset of model parameters \cite{38, 95, 111, 248, 333} or compressing gradients \cite{272, 270, 371}. One paper focused on training for several tasks simultaneously to avoid training independent models \cite{89}.\\

\noindent
\textbf{Component 4: Aggregation}\\

\noindent
\emph{Methodological advances.} In 37/89 papers, the authors focus on improving the aggregation component. Most commonly, papers consider the weighting of the contributions of the nodes in the network (11/37) by considering the local training loss \cite{389, 194}, local classification performance \cite{167}, using the signal-to-noise ratio of the data \cite{165}, based on node similarity \cite{202}, data quality \cite{192}, fairness \cite{9}, Shapley values \cite{136}, model performance on aggregator test data \cite{318} and by how each local model performs at all other nodes \cite{108}. 

Aggregation methods are often improved by measuring and correcting for the distribution differences between nodes \cite{29, 47, 143, 393}, allowing for multi-modal data aggregation across different nodes \cite{106, 69}. In 5/37 papers, the authors cluster nodes together based on their similarity before aggregation and 5/37 papers focus on improving the aggregation strategy. This is accomplished using hierarchical aggregation \cite{365}, performing some training rounds on the aggregator \cite{115}, dynamically selecting participating nodes and scheduling of aggregation \cite{10} and the ensembling of the local models to obtain a final model \cite{31, 137}. The remaining papers focus on methods for asynchronous aggregation (3/37) \cite{101, 185, 379}, using knowledge distillation (2/37) \cite{13, 30}, secure aggregation (2/37) \cite{155,355}, feature selection in a federated manner \cite{169}, attack detection during aggregation \cite{257} and aggregation of heterogeneous model architectures \cite{20}. \\

\noindent
\emph{Model aggregation techniques.}
The updates and contributions from nodes to improve the global model were mentioned in 78/89 papers. Most papers focused on improving the aggregation strategy, with 10/78 developing custom simple averaging strategies 
and 31/78 developing customised weighted averaging methods. %
The well-known FedAvg \cite{mcmahan2017communication} aggregation method was used in 14/78 papers %
with a modified formulation of it employed in a further 10/78. %
The remainder of the studies used non-averaging-based methods such as knowledge distillation \cite{13,30}, stacking \cite{20, 31}, split learning \cite{137}, feature fusion \cite{106}, Federated Goal Programming \cite{108}, and training the model one node at a time \cite{180}. See Table~\ref{Tab:1} for details.\\

\noindent
\emph{Synchronous vs. asynchronous updates.}
Only four studies \cite{101, 185, 222, 379} developed FL methods that allowed for asynchronous updates to the global model, whilst all others required the local nodes to finish their optimisation before the updates were applied. Asynchronous updates were made after comparing the performance of the model with the new and previous parameters and only updating the model if the update performed better.\\

\noindent
\emph{Node weighting.} 
A weighting factor was discussed in 50/89 papers to balance the node contributions during aggregation.
The majority of papers (38/50) %
weighted the contribution based on the sample size at each node. 
Two papers \cite{13,167} assigned weights based on the classification accuracy performance of each node, one weighted by considering the credibility of each node \cite{318}, and another dynamically adjusted the weights based on the variation of loss values from the previous round \cite{389}. Some papers did not aggregate the local contribution from a node if it was not beneficial to the overall network \cite{332, 185} or did not satisfy defined performance criteria \cite{167}. Model training time was considered in one paper \cite{10} to limit the number of devices transmitting their updates to the aggregator. A reward mechanism was used in one paper \cite{267} to select optimal nodes, considering the quality of their data and associated energy costs. One paper \cite{31} compared multiple means from which to determine the weighting during aggregation.\\

\noindent
\emph{Distribution comparisons.}
In order to address data heterogeneity between sites, only three papers \cite{381, 202, 375} mentioned that the model distributions between nodes were compared for clustering purposes before aggregation.\\

\noindent
\textbf{Component 5: Redistribution}\\

\noindent
\emph{Methodological advances.} In 4/89 papers the authors focus on improving the way models are redistributed to local nodes. Two papers personalise the model that is redistributed to the local nodes based on the similarity to other nodes in the network \cite{202} and a comparison of the gradient values against the global model \cite{381}. The other two papers focus on improving access to the global model by allowing local nodes or registered third-party researchers to run encrypted inference directly on the global model \cite{61, 76}. \\

\noindent
\emph{Training termination.} 
The criteria for ending model redistribution were specified in 79/89 papers. The majority (69/79) %
used a pre-determined number of update rounds before termination. Three papers \cite{20, 29, 373} required only a single round of communication by design. Two papers \cite{379, 393} used a minimum loss threshold to claim convergence of the global model and terminate re-distribution.\\

\noindent
\subsection{FL Evaluation Criteria}
\label{sec:flEvaluationCriteria}
\vspace{0.1in}

The papers we reviewed each focused on optimising the performance of the FL methods in several different domains, such as improved model efficacy, reduced communication overhead and efficient consumption of resources. \\

\noindent
\emph{Model efficacy.}
Common performance metrics were reported in most papers, such as the model accuracy (59/89), %
area under the receiver operator characteristic curve (17/89),  %
precision (15/89), %
F1-score (20/89), %
sensitivity/recall (18/89), %
specificity (6/89), %
Dice score (8/89) %
and loss value (12/89). %

There were 34 papers %
which benchmarked their proposed methods against other FL methods whilst 27 studies %
compared their simulated FL approaches with classic centralised ML. 
Given the focus of this review was on papers introducing novel contributions to the FL methodology, often ablation studies were performed to assess the impact on performance by including and excluding their proposed modifications \cite{10, 20, 30, 69, 77, 272, 379, 139, 180, 169, 202, 283, 265, 12}.
Seven papers compared the performance of different local model architectures while keeping the FL framework the same \cite{20, 106, 154, 169, 205, 265, 58}. Furthermore, 34 studies consider multiple hyperparameter configurations for their proposed adaptations. \\ %

\noindent
\emph{Communication efficiency.}
Key metrics for measuring the communication overhead were considered in several papers, such as communication cost (13/89) \cite{9, 16, 29, 60, 106, 111, 169, 192, 205, 270, 300, 332, 371}, number of communication rounds (4/89) \cite{111, 112, 300, 328} and the latency (4/89) \cite{106, 10, 101, 224}. \\

\noindent
\emph{Resource Consumption.}
One paper \cite{344} focused on CPU processing time, total training time, memory usage, and energy consumption for different numbers of offloaded layers in mobile devices. Time was an important consideration in several papers, including training time (5/89) \cite{101, 344, 373, 215, 169}, model parameter encryption time (2/89) \cite{61, 333}, and authentication signature compute time \cite{16}. Energy consumption was measured in four papers \cite{267, 344, 101, 215}.\\

\section{Discussion}

\noindent
Our systematic review has identified a keen appetite in the research community for developing FL methods with application to diverse healthcare problems. 
Most studies explore HFL, whilst VFL is under-explored in the literature. This is surprising as VFL holds immense potential for addressing healthcare problems, especially considering that healthcare data often has the inherent challenges of being siloed due to logistical issues and privacy concerns related to extensive data linking. We have identified many different systemic issues in the FL literature, which we will identify and then give recommendations for corrective action. FL methods have several distinct components and this review uniquely focuses on the methodological advances and challenges for each of them. \\

\noindent
\textbf{Reproducibility}\\

\noindent
\emph{Documentation issues.} In general, the documentation of papers was not sufficient to allow for the reproduction of the study findings. We found that key details such as the data pre-processing techniques, missing data imputation methods, model initialisation strategy, and optimisers employed were missing in a large proportion of the papers reviewed. These basic details are crucial to allow for the reproduction of the results that the papers describe. There is also poor documentation of the data exchanged from nodes to the aggregator, with the terms ``model parameters'' and ``model updates'' used interchangeably without specifying if these are the gradients, model weights or some other parameters. \\

\noindent
\emph{Codebase issues.} It is also very surprising that most papers develop their own implementation of the FL codebase rather than leveraging and building upon existing FL frameworks. The extreme complexity of FL systems leads to concerns that individual implementations are likely to suffer from issues of correctness unless carefully developed \cite{dittmer2023navigating}. Also, no papers released trained models and therefore, it is not possible to assess the performance of models independently. \\ 

\noindent
\emph{Recommendations for reproducibility.} As FL is a rapidly evolving field of innovative research, we recommend that the community work together to develop an FL methodology checklist to improve the documentation of future studies. In the absence of such a checklist, we recommend that authors and reviewers use existing checklists, such as CLAIM~\cite{mongan2020checklist}, for assessing the completeness of the data and model descriptions. Additionally, tools such as PROBAST~\cite{wolff2019probast} are recommended for assessing the biases in the data and models. Practitioners should only develop a new FL codebase when the existing frameworks fundamentally do not accomplish their aim, otherwise, there is a risk of coding errors due to the complexity of the FL system. Codebases and trained models should be released publicly if possible to allow the community to easily apply the model and validate the performance. \\

\noindent
\textbf{Datasets}\\

\noindent
\emph{Data missingness issues.} Given the nature of healthcare data, where different variables will be obtained at different sites, it is surprising that no studies discussed structural or informative missingness \cite{mitra2023learning,van2018flexible} with only two considering imputation of missing data. One paper even mentions missing values within their data without describing how the missingness was addressed. It has been shown in other studies that poor quality imputation, and imputation for non-random missingness, can bias a model trained using it \cite{groenwoldInformativeMissingnessElectronic2020, shadbahr2022classification}.\\ 

\noindent
\emph{Imaging data issues.} As clinical imaging studies often suffer from low patient numbers due to low disease prevalence at individual sites, FL gives a potential method to circumvent this issue \cite{rieke2020future} and we found 32/89 studies applied FL to imaging. However, given the known challenges of applying ML to imaging data \cite{roberts2021common, simko2022reproducibility}, FL has the potential to amplify these issues as the inherent data biases cannot be explored. \\

\noindent
\emph{Pre-processing issues.} Where pre-processing of data is utilised, it is always performed at local nodes or centrally before distribution to the nodes for artificial FL setups. This local approach, however, propagates any biases in feature values through to the pre-processed data.\\

\noindent
\emph{Data encryption issues}. Whilst many authors enhanced the privacy of the network by encrypting the exchanged data, all models were trained using pre-processed raw data. No paper considered a one-way hashing or encrypting of data before performing training \cite{lee2018privacy}. This is alarming, given that model weights can be highly informative about the raw data, whereas hashing of the data at the source allows for breakage of this link \cite{qiu2022all}.\\

\noindent
\emph{Class imbalance issues.} When addressing problems involving low disease prevalence, class imbalances are a reality and represent a critical issue for ML methods. These are compounded by FL methods, as there may be varying disease prevalence between nodes \cite{wang2021addressing}. However, this issue was only considered and addressed in six of the papers we reviewed.\\

\noindent
\emph{Node consistency issues.} Only one study checked for consistency in the data distributions between sites. In a real-world deployment scenario, biases between different sites are a reality that must be considered \cite{fang2022robust}. For example, in an FL network of hospitals, it would be important to understand whether data for paediatric or maternity hospitals were hosted at particular nodes to fully appreciate the age and sex biases inherent to the data.\\

\noindent
\emph{Consistent data filtering issues.} It is also important to note that only one paper discussed how the authors derived the particular cohort at each node for their analysis. For example, if each node has EHR data, then for each use case, the EHR at each node must be filtered with defined inclusion and exclusion criteria.\\

\noindent
\emph{Partitioning issues.} Only two papers employed an internal validation and holdout cohort and most papers considered either exclusively a validation cohort or a holdout cohort. In the wider ML literature, the use of validation data to avoid model overfit and a holdout cohort for evaluation is standard practice, so it is surprising that FL literature does not echo this.\\

\noindent
\emph{Recommendations for data.} 
The types of missingness found in datasets should be detailed and any approaches to imputation stated. 
Where imaging data is used, extreme care should be taken to ensure biases in the data are understood and the best practice is followed in the local optimisation by using checklists, e.g. CLAIM\cite{mongan2020checklist}.
For real-world deployment, it would be preferable for e.g. feature value normalisation to be based on globally exchanged parameters or transformations \cite{marchand2022securefedyj}. We would encourage authors to consider hashing their datasets at source before use in the training network to mitigate against data reconstruction attacks. 
Class imbalances across nodes and within datasets at nodes should be factored into the methodologies, as they are a primary source of bias for FL in healthcare, and details of the imbalances should be disclosed in manuscripts.
Clinical data can inherently contain large biases and issues from a range of sources, such as impossible values, missing values and corruption. In the absence of direct access to the data at each node, automated pipelines should be integrated that can perform a quality and integrity check and remove known sources of bias \cite{breger2023pipeline}. Securely confirming, across nodes, that the demographics are similar to one another, within a tolerance range, is also encouraged \cite{zhou2018statistical}. The cryptographical field of multi-party computation could also be used to obtain data metrics from multiple sites without the need to share the data directly.
For applying inclusion and exclusion criteria to EHR or tabular datasets at nodes, authors should disclose the filters used to curate the cohort used in the experiments with, e.g., an SQL query \cite{Kotechae069048}. We encourage more widespread adoption of the use of internal validation and holdout cohorts to ensure performance is not overstated. \\

\noindent
\textbf{Local optimisation}\\

\noindent
\emph{Computational requirements issues.} In most papers, local optimisation of the models required GPU compute capability linked to the data source, which is not currently found in most hospital environments. There is some progress towards this with the increasing adoption of cloud computing capabilities in clinical environments along with Trusted Research Environments \cite{kavianpour2022next}. However, the lack of widespread computational capability remains a barrier to the mass adoption of FL in real-world healthcare settings.\\

\noindent
\emph{Convergence issues.} For training of local models, a fixed number of iterations were employed by most papers, with only one mentioning early stopping and two others specifying ``until convergence''. This is highly irregular, with principled stopping criteria widespread in the non-FL ML literature, such as early stopping \cite{prechelt2002early}. \\

\noindent
\emph{Recommendations for local optimisation.} 
It is important for authors to consider whether their solutions are of practical use in the hospital environment, where computing capabilities are more limited than in simulated experiments. The required hardware, most importantly if a GPU is required, should be disclosed in the manuscripts and disclosed as a limitation if not widely available.
Outside of FL settings, local optimisation is best terminated when the loss converges for a validation cohort of the data. FL, in particular, would benefit significantly from early stopping as the local node compute times can be highly variable, depending on the sample number and model architecture at each site.
As some studies have done, we recommend attempting to balance these local training hyperparameters depending on the local compute or data availability.\\

\noindent
\textbf{Communication}\\

\noindent
\emph{Metadata communication issues.} It was identified in several papers that metadata, beyond the model parameters, was often communicated in parallel to the aggregator. For instance, nearly all papers that relied on FedAvg as an aggregation method required the sharing of the sample number at each node. This can severely compromise the security of the network if an attacker intercepts the communication or compromises the aggregator by highlighting to a hostile actor those nodes that contain a large amount of data.\\

\noindent
\emph{Encryption issues.} Given that all of the papers focused on applications to healthcare, where privacy is a primary concern, it was surprising that most papers developed FL networks that do not encrypt the model parameters when being exchanged with the aggregator. It has been shown that private information contained within the training data can leak into the learned parameters of a model \cite{NEURIPS2020_c4ede56b, zhangBroadeningDifferentialPrivacy2020, fredriksonModelInversionAttacks2015}.\\

\noindent
\emph{Authentication issues.} FL networks are susceptible to attacks by a node exchanging inauthentic or false weights or by allowing the network to train on poisoned data \cite{pmlr-v108-bagdasaryan20a}. Only two papers authenticated nodes before accepting their update parameters.\\

\noindent
\emph{Recommendations for communication.}
It would be preferable for the number of samples at each node to never be shared with the aggregator as this information could be useful to a hostile actor and consideration in the literature should be given to aggregation methods that do not require the sample numbers. In the context of healthcare data, FL networks should use an established cryptography technique \cite{ma2020safeguarding} to encrypt as much of the data and communication as possible. Authors must decide and justify in the manuscripts the level of privacy required and balance this with the increased communication overhead. It is incredibly important for the aggregator to authenticate and authorise all nodes which data is accepted from and to which data are redistributed. Without this, the network is vulnerable to many different attacks \cite{reina2021openfl, li2023catfl}.\\

\noindent
\textbf{Aggregation}\\

\noindent
\emph{Aggregation issues.} Most papers required the use of an aggregator, with only eight studies exploring a fully decentralised architecture. Decentralisation allows for a network that is more robust to failure and attack \cite{kairouz2021advances} with updates performed mutually between the nodes themselves. Only three papers, discussed a method that allows for asynchronous updates to the global model and almost all aggregation techniques are simple or weighted averaging of the contributions from each node. Requiring synchronous updates is a limitation for real-world deployment as differing local optimisation speeds will result in nodes lying idle, waiting for the aggregator to return the global model only once all nodes have finished training.\\

\noindent
\emph{Update comparison issues}. Nodes may return very different updates to the global model for incorporation at the aggregation stage. Most of the papers which disclose their method, simply weight the contributions by sample size alone. However, it is useful to understand whether those updates are consistent with one another or whether one or more nodes are suggesting updates that are vastly different from the others, as this may be symptomatic of issues with the data or training \cite{cho2022towards, abay2020mitigating}. Two papers have introduced methods that allow for the ignoring or minimisation of the contributions of those nodes which are not beneficial to the overall network. \\

\noindent
\emph{Recommendations for aggregation.} 
Authors should consider whether synchronous or asynchronous updates are preferable for their use case, especially if the local optimisations have very different training times or not all nodes are always available. When the nodes are supplying updates to the global model, they should be assessed for consistency with other nodes. If a node gives very different model updates compared to the others, this should be investigated.
Of the papers we reviewed, none performed such comparisons.\\

\noindent
\textbf{Redistribution}\\

\noindent
\emph{Convergence issues.} The global redistribution was always performed for a fixed number of epochs rather than a principled stopping criterion. This is aberrant compared to the wider ML literature, where early-stopping and validation data are used to terminate training. \\

\noindent
\emph{Recommendations for redistribution.} 
Communication rounds should be terminated in a principled way. This could be based on the performance of the global model at each node, on a validation or holdout cohort, or the performance of the global model on evaluation data held at the central aggregator.\\

\noindent
\textbf{Deployment}\\

\noindent
\emph{Deployment issues.} 
None of the papers provide evidence about the deployment of their FL platform in a healthcare environment. It needs to be clarified how each node is set up in individual hospitals, how the local model is delivered to these nodes, and how these nodes connect to existing databases. The communication protocols enabling interactions between nodes and aggregators were largely unspecified. Moreover, the mechanism or event that triggers new training rounds was not described.\\

\noindent
\emph{Recommendations for deployment.} 
For FL algorithms applied to healthcare, it is crucial that deployment is carefully considered and planned for. This strategy should consider the entire process, from node distribution to aggregator interaction, ensuring seamless communication and efficient training rounds. Furthermore, the authors should consider integrating MLOps practices, as these enhance automation, monitoring, and security, ensure seamless integration and deployment, encourage collaboration, and increase the efficiency and reliability of the FL platform. All papers we reviewed did not discuss version control of the global model. Implementing version control in FL enhances traceability, supports asynchronous communication, enables A/B testing, and provides a rollback mechanism. Storing copies of model artifacts across different versions strengthens auditability and facilitates benchmarking and comparing model performance over time.\\

\noindent
\textbf{Limitations.} The scope of this review leads it to focus only on those papers whose contribution is to the methodology of FL applied to healthcare. We do not consider manuscripts that apply off-the-shelf FL methods directly to the data. This is a limitation to the review as many manuscripts have not been considered that claim to have been successfully applied to healthcare data. Additionally, we have not assessed manuscript quality using checklists or performed a review of bias (ROB) assessment. Currently, we have been unable to identify an approach appropriate for FL papers and it is unclear how to fairly generalise checklists such as CLAIM \cite{mongan2020checklist} or ROB frameworks such as PROBAST \cite{wolff2019probast} to an FL setting.\\

\section{Conclusions}

This review focused on the literature describing FL methods for healthcare applications where there were methodological advances. We considered the different areas in which methodological advancements are being made whilst systematically exploring the application areas and how the FL components were being developed. In each component, we identified systemic pitfalls and gave recommendations to support those who are developing FL in healthcare. Specifically, there are significant improvements required in areas such as documentation quality, addressing imbalanced and missing data and sharing of non-encrypted updates. The community must also work together to design appropriate checklists for FL methods in healthcare and review of bias frameworks for this setting. FL will become a more common and significant tool for healthcare analytics in the future, and by following these best practice recommendations, we increase the likelihood of the adoption of these tools in clinical practice.

\section*{Methods}\label{sec:methods}

\textbf{Review strategy and selection criteria.} We performed a search of published works using Scopus for phrases shown in the `Search terms' section covering the period from January 2015 to February 2023. The review was performed using the Covidence\cite{covidence} systematic review platform. \\

\noindent
\emph{Search terms.} An initial search was performed to extract papers containing one of "federated learning" or "distributed learning" along with one of "classify", "predict", "prediction", "identify", "predictive", "prognosticate", "diagnosis", "diagnostic", "diagnose", "outlier", "anomaly", "detect", "detecting" in the title or abstract. We also require that "healthcare", "health" or "health care" appear in the abstract. We excluded articles focussing on blockchain development, intrusion detection, remote teaching and systematic reviews by excluding those papers which include "blockchain", "block chain", "classroom", "class room", "attack" in the title or abstract or have "intrusion", "intrude", "review", "survey" in the title. \\

\noindent
\emph{Title and abstract screening.} A team of eight reviewers screened the titles and abstracts for each of the papers. Each paper was independently assessed by two reviewers and conflicts were resolved by consensus of all reviewers. \\

\noindent
\emph{Full-text screening.} Nine reviewers performed the full-text screening, with each paper independently assessed by two reviewers. Any conflicts were resolved by the consensus of all reviewers. \\

\noindent
\emph{Data extraction.} A team of four extracted the data from each paper that was used to write the manuscript and assemble Table~\ref{Tab:1}. \\

\noindent
\textbf{Role of the funding source.} The funders of the study had no role in the study design, data collection, data analysis, data interpretation or the writing of the manuscript. All authors had full access to all the data in the study and had final responsibility for the decision to submit for publication. \\

\noindent
\textbf{Conflicts of interest.} The authors declare no conflicts of interest.\\

\noindent
\textbf{Acknowledgements}
The authors are grateful for the following funding: The Trinity Challenge (F.Z., D.K., S.T., BloodCounts! Collaboration, C.-B.S., N.G., M.R.), the EU/EFPIA Innovative Medicines Initiative project DRAGON (101005122) (T.S., C.-B.S., M.R.), AstraZeneca (Y.C.), the EPSRC Cambridge Mathematics of Information in Healthcare Hub EP/T017961/1 (J.H.F.R., J.A.D.A, C.-B.S., M.R.), the Cantab Capital Institute for the Mathematics of Information (C.-B.S.), the European Research Council under the European Union’s Horizon 2020 research and innovation programme grant agreement no. 777826 (C.-B.S.), the Alan Turing Institute (C.-B.S.), Wellcome Trust (J.H.F.R.), Cancer Research UK Cambridge Centre (C9685/A25177) (C.-B.S.), British Heart Foundation (J.H.F.R.), the NIHR Cambridge Biomedical Research Centre (J.H.F.R.), HEFCE (J.H.F.R.).
In addition, C.-B.S. acknowledges support from the Leverhulme Trust project on ‘Breaking the non-convexity barrier’, the Philip Leverhulme Prize, the EPSRC grants EP/S026045/1 and EP/T003553/1 and the Wellcome Innovator Award RG98755.\\

\noindent
\textbf{BloodCounts! Consortium}
Martijn Schut$^{9}$, Folkert Asselbergs$^{9}$, Sujoy Kar$^{10}$, Suthesh Sivapalaratnam$^{11}$, Sophie Williams$^{11}$, Mickey Koh$^{12}$, Yvonne Henskens$^{13}$, Bart de Wit$^{13}$, Umberto D'Alessandro$^{14}$, Ute Jentsch$^{15}$, Parashkev Nachev$^{16}$, Rajeev Gupta$^{16}$, Sara Trompeter$^{16}$, Nancy Boeckx$^{17}$, Christine van Laer$^{17}$, Gordon. A. Awandare$^{18}$, Lucas Amenga-Etego$^{18}$, Mathie Leers$^{19}$, Mirelle Huijskens$^{19}$, Samuel McDermott$^{1}$, James Rudd$^{6}$, Carola-Bibiane Schönlieb$^{1}$, Nicholas Gleadall$^{8}$ and Michael Roberts$^{1,6}$.\\ 

\noindent
${}^{9}$ Amsterdam University Medical Centre, Amsterdam, Netherlands.
${}^{10}$ Apollo Hospitals, Chennai, India.
${}^{11}$ Barts Health NHS Trust, London, United Kingdom.
${}^{12}$ Health Services Authority, Singapore.
${}^{13}$ Maastricht University Medical Centre, Maastricht, Netherlands.
${}^{14}$ MRC The Gambia Unit, Banjul, The Gambia.
${}^{15}$ South African National Blood Service, Gauteng, South Africa.
${}^{16}$ University College London Hospitals, London, United Kingdom.
${}^{17}$ University Hospitals Leuven, Leuven, Belgium.
${}^{18}$ West African Centre for Cell Biology of Infectious Pathogens, Accra, Ghana.
${}^{19}$ Zuyderland Medical Center, Zuyderland, Netherlands.

\end{multicols}

\begin{landscape}

\setlength\LTleft{0pt}
\setlength\LTright{0pt}
\begin{longtable}{|l|p{0.3in} | p{0.5in}| p{0.4in}|p{0.4in}|p{0.65in}|p{0.5in}|p{0.4in}|p{0.4in}|p{0.5in}|p{0.5in}|p{0.5in}|p{0.6in}}
\caption {Data extracted for reviewed papers}\label{Tab:1}\\\toprule
\endfirsthead
\caption* {\textbf{Table \ref{Tab:1} Continued:} Data extracted}\\\toprule
\endhead
\endfoot
\bottomrule
\endlastfoot
\toprule
Authors & FL Type & Applic. & Data Type & Pre-process. & Model Type & Optimiser & Init. & Local Training Termination & Data Exch. & Exch. Encrypted & Agg. Method \\
\midrule
Akter M. et al. \cite{27} & HFL & Class. & Imaging & \checkmark & CNN & --- & PD & Fixed & Model Params. & \xmark & FedAvg \\
Alam M.U. et al. \cite{12} & HFL & Class. & EHR & \checkmark & RNN-LSTM & Adam & Rand. & Fixed & Model Params. & \xmark & W. Avg. \& FedOpt \\
Aminifar A. et al. \cite{224} & HFL & Class. & Sensor & \xmark & Rand. Forests & --- & Rand. & Conv. & N/A & \checkmark & --- \\
Balkus S.V. et al. \cite{112} & HFL & Cluster. & Diet. Study & \checkmark & Fuzzy Cluster. & --- & PD & Fixed & Model Params. & \xmark & Simple Avg. \\
Bey R. et al. \cite{343} & HFL & Regress. & EHR & \checkmark & Gradient-Boosted Trees & --- & --- & Fixed & N/A & \xmark & --- \\
Brisimi T. et al. \cite{300} & HFL & Class. & EHR & \checkmark & SVM & --- & PD & Fixed & Model Params. & \xmark & W. Avg. \\
Camajori Tedeschini B. et al. \cite{222} & HFL & Class. \& Segment. & Imaging & \checkmark & CNN & Adam & PD & Fixed & Weights \& Model Hyperparameters & \checkmark & W. Avg. \\
Cetinkaya A.E. et al. \cite{377} & HFL & Class. & Imaging & \xmark & CNN & Adadelta & Rand. & Fixed & Weights & \xmark & FedAvg \\
Chang C. et al. \cite{143} & HFL & Regress. & EHR & \checkmark & Linear Regress. & MLE & --- & Conv. & MLE \& Posterior Samples & \xmark & Custom Alg. \\
Che S. et al. \cite{69} & HFL \& VFL & Class. & Sensor & \checkmark & RNN & RMSprop & Rand. & Fixed & Model Params. & \xmark & W. Avg. \\
Chen H. et al. \cite{332} & HFL & Class. & Non-medical & \xmark & CNN & --- & Rand. & Fixed & Gradients & \checkmark & Simple Avg. \\
Chen Y. et al. \cite{318} & HFL & Class. & Sensor & \checkmark & CNN & SGD & Rand. & Fixed & Gradients & \xmark & Custom. FedAvg \\
Cholakoska A. et al. \cite{243} & HFL & Anomaly Detection & N/A & \xmark & --- & --- & --- & Fixed & Model Params. & \xmark & --- \\
Choudhury O. et al. \cite{346} & HFL & Class. & EHR & \checkmark & SVM \& Log. Regress. & --- & --- & --- & Weights & \xmark & --- \\
Chu D. et al. \cite{111} & FTL & Class. & Sensor & \checkmark & CNN & Adam & --- & Fixed & Weights & \xmark & W. Avg. \\
Feng T. et al. \cite{154} & HFL & Class. & Speech & \checkmark & MLP & --- & Rand. & Fixed & Model Params. & \xmark & Simple Avg. \\
Foley P. et al. \cite{35} & HFL & N/A & N/A & \xmark & --- & --- & --- & --- & Weights & \checkmark & --- \\
Gad G. et al. \cite{13} & HFL & Class. & Sensor & \checkmark & CNN \& RNN & SGD \& Adam \& RMSprop & Rand. & Fixed & Output & \xmark & W. Avg. \\
Gencturk M. et al. \cite{167} & HFL & Class. & Imaging & \xmark & Rand. Forests & Custom. & Rand. & Fixed & Dec. Trees & \xmark & W. Avg. \\
Gong Q. et al. \cite{64} & HFL & Class. & Sensor & \checkmark & RNN & --- & Rand. & Conv. & Model Params. & \xmark & FedAvg \\
Guo K. et al. \cite{191} & HFL & Class. & Imaging & \xmark & CNN & SGD & Rand. & Fixed & Weights & \xmark & FedAvg \\
Guo Y. et al. \cite{344} & HFL & Class. & Imaging & \xmark & MLP & --- & Rand. & Fixed & Weights & \xmark & Simple Avg. \\
Gupta D. et al. \cite{365} & HFL & Anomaly Detection & EHR & \xmark & RNN & --- & Rand. & Fixed & Weights & \xmark & W. Avg. \\
Hao M. et al. \cite{333} & HFL & Class. & Non-medical & \checkmark & CNN & SGD & Rand. & Fixed & Gradients & \checkmark & Simple Avg. \\
Hilberger H. et al. \cite{77} & HFL & Class. & Imaging & \checkmark & MLP & Adam & PD & Fixed & Weights & \xmark & FedAvg \\
Ji J. et al. \cite{106} & VFL & Class. & Sensor & \xmark & CNN & Adam & --- & --- & Output & \checkmark & Feature Fusion \\
Kalapaaking A.P. et al. \cite{61} & HFL & Class. & Imaging & \xmark & CNN & --- & Rand. & Fixed & Weights & \checkmark & FedAvg \\
Kandati D.R. et al. \cite{51} & HFL & Class. & EHR & \xmark & MLP & Adam & Rand. & Fixed & Weights & \xmark & Simple Avg. \\
Kerkouche R. et al. \cite{272} & HFL & Class. & EHR & \checkmark & MLP & SGD & Rand. & Fixed & Model Params. & \xmark & W. Avg. \\
Kim J. et al. \cite{393} & VFL & Class. & Genom. & \xmark & Log. Regress. & Netwon's method & PD & Fixed & Gram Matrix & \xmark & Custom Alg. \\
Kumar S. et al. \cite{136} & HFL & Class. & Imaging & \xmark & CNN & SAGA & Rand. & Fixed & Weights & \xmark & Custom. FedAvg \\
Li J. et al. \cite{185} & HFL & Class. & Non-medical & \xmark & CNN & SGD & Pre-train. \& Rand. & Fixed & Weights & \checkmark & Asynch. FedAvg \\
Li J. et al. \cite{97} & HFL & Class. & Speech & \xmark & Log. Regress. & SGD & --- & Fixed & Weights & \checkmark & FedAvg \\
Li L. et al. \cite{142} & HFL & Class. & Sensor & \xmark & CNN & SGD & Rand. & Fixed & Model Params. & \xmark & W. Avg. \\
Li Z. et al. \cite{194} & HFL & Class. & Imaging & \xmark & CNN & SGD & PD & Fixed & Weights \& Loss & \xmark & W. Avg. \\
Lian Z. et al. \cite{205} & HFL & Class. & Imaging & \checkmark & CNN & SGD & Pre-train. & Fixed & Weights & \xmark & W. Avg. \\
Liu X. et al. \cite{165} & HFL & Regress. & Video & \checkmark & TS-CAN & Adam & Rand. & Fixed & Weights \& Signal Quality Score & \xmark & Custom. FedAvg \\
Liu X. et al. \cite{58} & HFL & Class. \& Architecture Search & Imaging & \xmark & CNN & SGD & --- & Fixed & Model Params. & \xmark & FedAvg \\
Lu S. et al. \cite{328} & HFL & N/A & EHR & \xmark & MLP & SGD & Rand. & Fixed & Gradients & \xmark & W. Avg. \\
Lu W. et al. \cite{202} & HFL & Class. & Imaging & \checkmark & CNN & SGD & Pre-train. & Fixed & Model Params. \& Batch-Norm Stats. & \xmark & W. Avg. \\
Ma J. et al. \cite{270} & HFL & Tens. Fact. & EHR & \xmark & Tens. Fact. & SGD & --- & Fixed & Gradients & \xmark & Simple Avg. \\
Ma J. et al. \cite{371} & HFL & Tens. Fact. & EHR & \xmark & Tens. Fact. & SGD & Rand. & Fixed & Factor Mats. & \xmark & CiderTF \\
Malik H. et al. \cite{10} & HFL & Class. & Imaging & \checkmark & CNN & --- & --- & Fixed & Gradients & \xmark & --- \\
Maryam Hosseini S. et al. \cite{9} & HFL & Class. & Imaging & \checkmark & CNN & SGD & Rand. & Fixed & Gradients & \xmark & FedSGD \\
Mathieu Andreux et al. \cite{416} & HFL & Class. \& Segment. & Imaging & \checkmark & CNN & Adam & Rand. & Fixed & Model Params. & \xmark & Custom. FedAvg \\
Mocanu I. et al. \cite{375} & HFL & Class. & Imaging & \checkmark & CNN & SGD & --- & --- & Weights & \xmark & FedAvg \\
Nguyen T.V. et al. \cite{30} & HFL & Class. & Imaging & \checkmark & CNN & SGD & Pre-train. & Fixed & Output & \xmark & Knowl. Dist. \\
Oh H. et al. \cite{213} & HFL & Class. & EHR & \checkmark & MLP & Adam & PD & Fixed & Weights & \xmark & W. Avg. \\
Papadopoulos P. et al. \cite{265} & HFL & N/A & N/A & \xmark & --- & --- & --- & --- & N/A & \checkmark & --- \\
Paragliola G. et al. \cite{38} & HFL & Class. & Sensor & \xmark & CNN \& RNN & --- & Rand. & Fixed & Weights & \xmark & Custom. FedAvg \\
Paragliola G. et al. \cite{95} & HFL & Class. & Sensor & \checkmark & CNN \& RNN & SGD & Rand. & Fixed & Weights & \xmark & Custom. FedAvg \\
Park S. et al. \cite{139} & HFL & Class. \& Segment. & Imaging & \xmark & CNN & SGD \& Adam & Pre-train. & Fixed & Gradients & \xmark & Simple Avg. \\
Park S. et al. \cite{283} & HFL & Class. \& Segment. \& Obj. Detect. & Imaging & \xmark & Transformer & SGD & Pre-train. & Fixed & Weights & \xmark & W. Avg. \\
Presotto R. et al. \cite{206} & FTL & Class. & Sensor & \xmark & MLP & Adam & Rand. & Fixed & Weights & \xmark & FedAvg \\
Qu L. et al. \cite{83} & HFL & Class. \& Regress. & Imaging & \checkmark & CNN & SGD & --- & Fixed & Latent Vars. \& Labels & \xmark & --- \\
Rajotte J.F. et al. \cite{245} & HFL & Class. & Imaging & \xmark & CNN & --- & Rand. & Fixed & N/A & \xmark & --- \\
Raza A. et al. \cite{107} & FTL & Class. & Sensor & \checkmark & CNN & RMSprop & PD & Fixed & Weights & \xmark & W. Avg. \\
Repetto M. et al. \cite{108} & HFL & Class. & EHR & \xmark & Gradient-Boosted Trees \& Log. Regress. \& MLP, Rand. Forest \& SVM & --- & --- & --- & Output & \xmark & Federated Goal Programming \\
Reps J.M. et al. \cite{31} & HFL & Class. & EHR & \checkmark & Log. Regress. & --- & --- & --- & Output & \xmark & W. Avg. \\
Roth H.R. et al. \cite{137} & VFL & Segment. & Imaging & \xmark & CNN & Adam & Rand. & --- & Gradients & \xmark & Split Learn. \\
Sakib S. et al. \cite{379} & HFL & Class. & Sensor & \xmark & CNN & --- & Pre-train. & Fixed & Weights & \xmark & --- \\
Sav S. et al. \cite{76} & HFL & Class. & EHR & \checkmark & CNN & SGD & --- & Fixed & Gradients & \checkmark & FedAvg \\
Shaik T. et al. \cite{20} & HFL & Class. & Sensor & \checkmark & CNN \& RNN & Adam & --- & --- & Output & \xmark & Stacking \\
Shen C. et al. \cite{389} & HFL & Segment. & Imaging & \xmark & CNN & Adam & Rand. & Fixed & Model Params. & \xmark & W. Avg. \\
Shen Z. et al. \cite{115} & HFL & Class. & Imaging & \checkmark & CNN & Adam & --- & Fixed & Model Params. & \xmark & W. Avg. \\
Shin Y.A. et al. \cite{155} & HFL & N/A & N/A & \xmark & --- & SGD & Rand. & Fixed & Weights & \checkmark & Custom. FedAvg \\
Souza R. et al. \cite{180} & HFL & Segment. & Imaging & \xmark & CNN & --- & PD & One Epoch & Weights & \xmark & Travel. Model \\
Sun W. et al. \cite{373} & VFL & Class. & Sensor & \xmark & ELM  & --- & Rand. & --- & Weights & \xmark & ELM \\
Thakur A. et al. \cite{89} & HFL & Class. & EHR & \checkmark & RNN & SGD \& Adam & PD & Fixed & Model Params. & \xmark & Simple Avg. \\
Tong J. et al. \cite{29} & HFL & Regress. & Claims & \xmark & Log. Regress. & --- & --- & --- & Gradients & \xmark & W. Avg. \\
Wang Q. et al. \cite{221} & HFL & Class. & Genom. & \xmark & --- & --- & Rand. & Fixed & Weights & \checkmark & W. Avg. \\
Wang R. et al. \cite{215} & HFL & Class. & Non-medical & \xmark & CNN & SGD & PD & Fixed & Weights & \checkmark & W. Avg. \\
Wang W. et al. \cite{16} & HFL & N/A & N/A & \checkmark & --- & --- & Rand. & Conv. & Model Params. & \xmark & W. Avg. \\
Wang X. et al. \cite{47} & FTL & Regress. & EHR & \xmark & Cox & --- & Rand. & --- & Model Params. & \xmark & W. Avg. \\
Wen T. et al. \cite{150} & HFL & Class. & Imaging & \checkmark & CNN & SGD & Rand. & Fixed & Model Params. & \xmark & Custom. FedAvg \\
Wu Q. et al. \cite{60} & HFL & Class. & Sensor & \checkmark & CNN & SGD & Rand. & Fixed & Weights & \checkmark & FedAvg \\
Wu Y. et al. \cite{378} & HFL & Class. & Imaging & \checkmark & CNN & Adam & Rand. & Fixed & Weights & \xmark & FedAvg \\
Xi B. et al. \cite{257} & HFL & Class. & EHR & \xmark & CNN & --- & Rand. & --- & N/A & \xmark & --- \\
Xie Y. et al. \cite{355} & HFL & Class. & N/A & \xmark & Rand. Forests & --- & --- & --- & Encryption Gradients & \checkmark & Ensemble \\
Yang Q. et al. \cite{248} & HFL & Class. & Imaging & \xmark & CNN & Adam & Rand. & Fixed & Gradients & \xmark & Simple Avg. \\
Yoo J.H. et al. \cite{381} & HFL & Class. & Sensor & \checkmark & CNN & SGD & Rand. & Fixed & Weights & \xmark & Custom. FedAvg \\
Yu C. et al. \cite{170} & HFL \& VFL & Class. & EHR & \checkmark & RNN & SGD & PD & Fixed & Output & \xmark & W. Avg. \\
Yu H. et al. \cite{385} & HFL & Class. & Sensor & \xmark & Transformer & Adam & Rand. & Fixed & Gradients & \xmark & W. Avg. \\
Zhang D.K. et al. \cite{14} & HFL & Regress. & EHR & \xmark & MLP & SGD & PD & Fixed & Model Params. & \xmark & W. Avg. \\
Zhang D.Y. et al. \cite{267} & HFL & Class. & Sensor & \xmark & --- & SGD & PD & Fixed & Weights & \xmark & Custom. FedAvg \\
Zhang L. et al. \cite{192} & HFL & Class. & Imaging & \checkmark & CNN & SGD & Rand. & Fixed & Model Params. \& Data Quality Metric & \checkmark & W. Avg. \\
Zhang R. et al. \cite{169} & VFL & Feat. Select. & EHR & \xmark & --- & --- & PD & --- & Binary Mats. & \checkmark & Custom Alg. \\
Zhang Y. et al. \cite{101} & HFL & Class. & Sensor & \xmark & CNN & SGD & Rand. & Fixed & Gradients & \xmark & W. Avg. \\
Zheng X. et al. \cite{391} & HFL & Class. & Sensor & \xmark & CNN & SGD & PD & Conv. & Weights & \checkmark & FedAvg
\\ \bottomrule
\end{longtable}

\end{landscape}

\bibliography{references}

\end{document}